\documentclass[10pt,twocolumn,letterpaper]{article}

\usepackage{cvpr}
\usepackage{times}
\usepackage{epsfig}
\usepackage{graphicx}
\usepackage{amsmath}
\usepackage{amssymb}
\usepackage{subfig}
\usepackage{cite}
\usepackage{caption}
\usepackage{xcolor}

\newif\ifsqueeze
\squeezetrue  % set \squeezetrue or \squeezefalse to enable or disable squeezing
\ifsqueeze
  \newcommand{\Caption}[1]{\vspace{-2.5mm}\caption{{\footnotesize #1}}}
  \newcommand{\Section}[1]{\vspace{-1mm} \section{#1} \vspace{-2mm}}
  
  \newcommand{\Subsection}[1]{\vspace{-1mm} \subsection{#1} \vspace{-1mm} }

\else
  \newcommand{\Caption}[1]{\caption{#1}}
  \newcommand{\Section}[1]{\section{#1}}
  
  \newcommand{\Subsection}[1]{\subsection{#1}}

\fi

\usepackage[pagebackref=true,breaklinks=true,letterpaper=true,colorlinks,bookmarks=false]{hyperref}

\newcommand{\denselist}{
\itemsep -2pt\topsep-8pt\partopsep-8pt
}

\cvprfinalcopy % *** Uncomment this line for the final submission

\def\cvprPaperID{7393} % *** Enter the CVPR Paper ID here

\ifcvprfinal\fi
\begin{document}

%%%%%%%%% TITLE
\title{Context R-CNN: Long Term Temporal Context for Per-Camera Object Detection}

\maketitle
%\thispagestyle{empty}
%%%%%%%%% ABSTRACT
\begin{abstract}
In static monitoring cameras, useful contextual information can stretch far beyond the few seconds typical video understanding models might see: subjects may exhibit similar behavior over multiple days, and background objects remain static. Due to power and storage constraints, sampling frequencies are low, often no faster than one frame per second, and sometimes are irregular due to the use of a motion trigger. In order to perform well in this setting, models must be robust to irregular sampling rates.
In this paper we propose a method that leverages
temporal context from the unlabeled frames of a novel camera to improve performance at that camera. Specifically,
we propose an attention-based approach that allows
our model, \textbf{Context R-CNN}, to index into a long term memory bank constructed on a per-camera basis and aggregate contextual features from other frames to boost object detection performance on the current frame.

We apply Context R-CNN to two settings: (1) species detection using camera traps, and (2) vehicle detection in traffic cameras, 
showing in both settings that Context R-CNN leads to performance gains over strong baselines.
Moreover, we show that increasing the contextual time horizon
leads to improved results.
When applied to camera trap data from the Snapshot Serengeti dataset, Context R-CNN with context from up to a \textbf{month} of images outperforms a single-frame baseline by
$17.9$\% mAP, and outperforms S3D (a $3$d convolution based baseline) by $11.2$\% mAP.  
\end{abstract}

%%%%%%%%% BODY TEXT
\vspace{-5pt}
\Section{Introduction}

\begin{figure}[h]
\centering
\includegraphics[width=8cm]{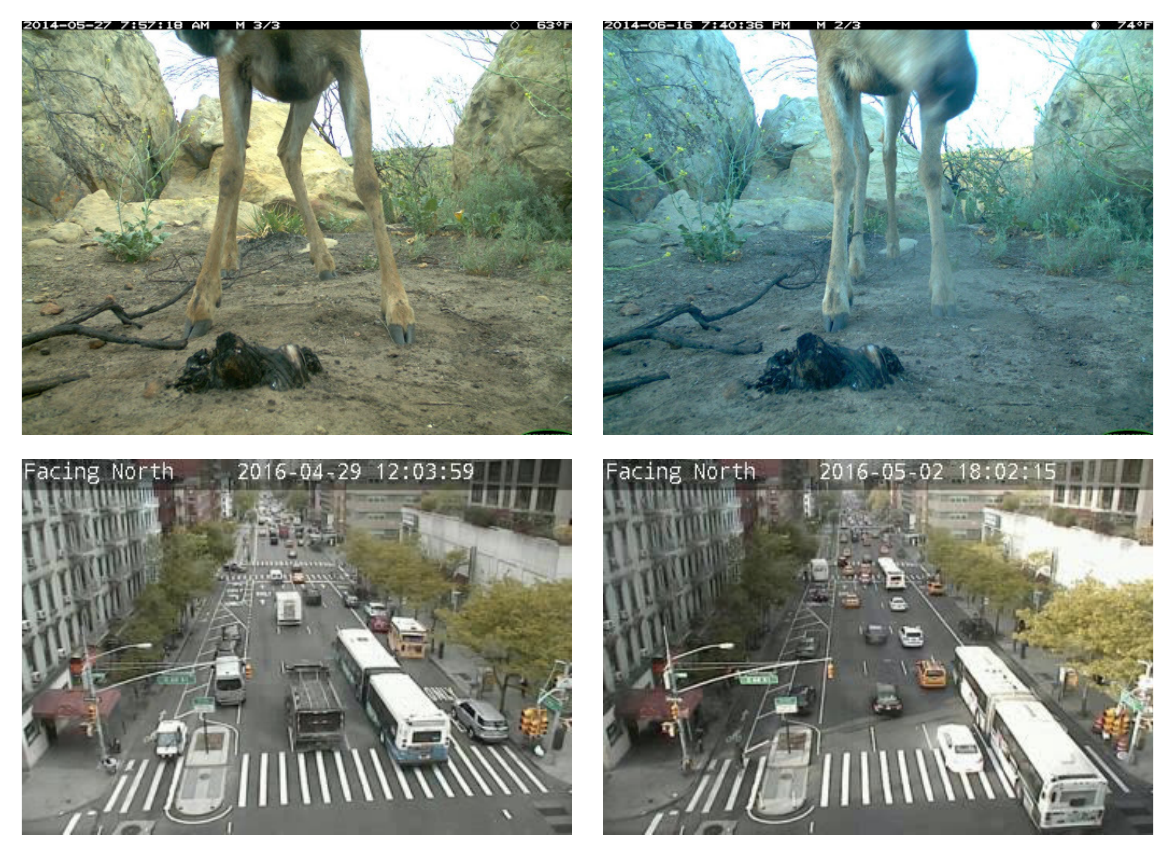}

\Caption{\textbf{Visual similarity over long time horizons.} In static cameras, there exists significantly more long term temporal consistency than in data from moving cameras. In each case above, the images were taken on separate days, yet look strikingly similar.}
\label{fig:image_similarity}
\end{figure}

We seek to improve recognition within passive monitoring cameras, which are static and collect sparse data over long time horizons.\footnote{Models and code will be released online.}  Passive monitoring deployments are ubiquitous and
present unique challenges for computer vision but also offer unique opportunities that can be leveraged for improved accuracy.

For example, depending on the triggering mechanism and the camera placement, large numbers of photos at any given camera location can be empty of any objects of interest (up to $75$\% for some ecological camera trap datasets) \cite{norouzzadeh2018automatically}. Further, as the images in static passive-monitoring cameras are taken automatically (without a human photographer), there is no guarantee that the objects of interest will be centered, focused, well-lit, or an appropriate scale. We break these challenges into three categories, each of which can cause failures in single-frame detection networks:

\begin{itemize}\denselist
    \item \textbf{Objects of interest partially observed.} Objects can be very close to the camera and occluded by the edges of the frame, partially hidden in the environment due to camouflage, or very far from the camera.
    \item \textbf{Poor image quality.} Objects are poorly lit, blurry, or obscured by weather conditions like snow or fog.
    \item \textbf{Background distractors.} When moving to a new camera location, there can exist salient background objects that cause repeated false positives. 
\end{itemize}

\begin{figure}[ht!]
  \centering

\subfloat[\label{fig:citycam_object_moving_out}][Object moving out of frame.]{
\includegraphics[width=8cm, height=2.65cm]{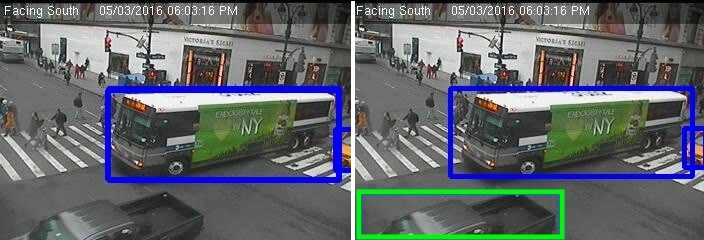}
}\vspace{-10pt} \\

\subfloat[\label{fig:salient_object}][Object highly occluded.]{\includegraphics[width=8cm,height=2.65cm]{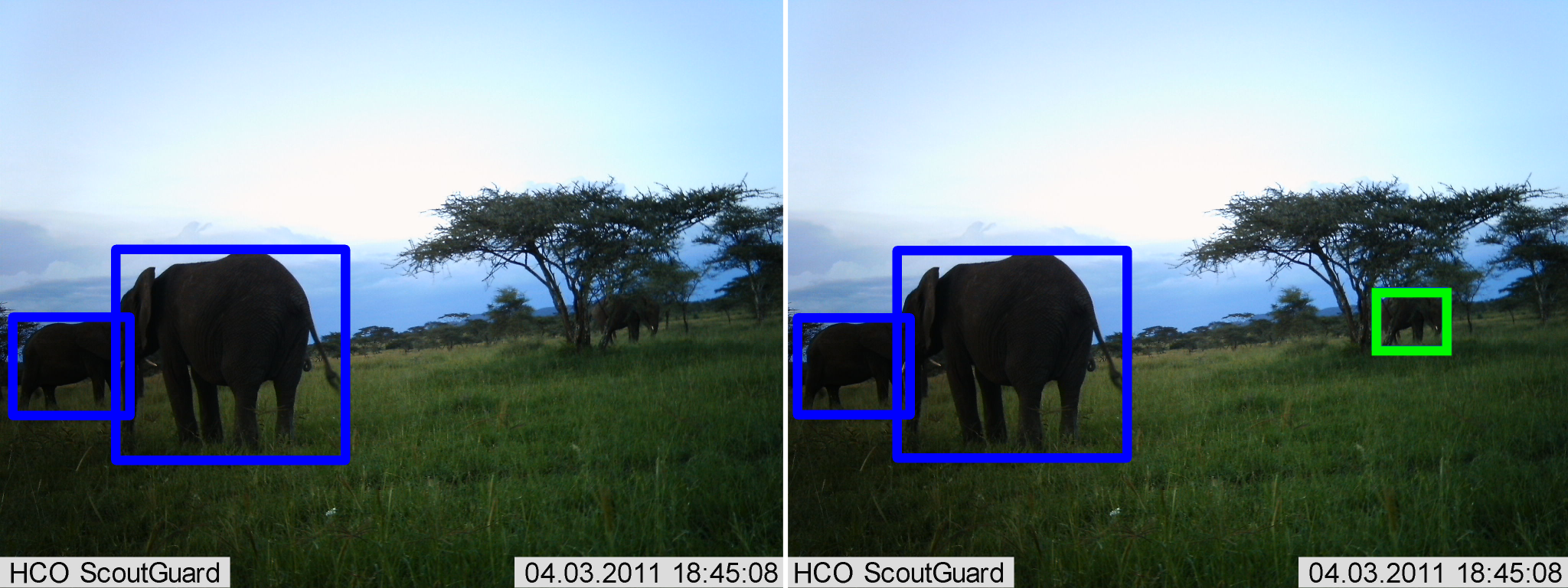}}
\vspace{-10pt} \\

\subfloat[\label{fig:citycam_object_moving_out}][Object far from camera.]{\includegraphics[width=8cm,height=2.65cm]{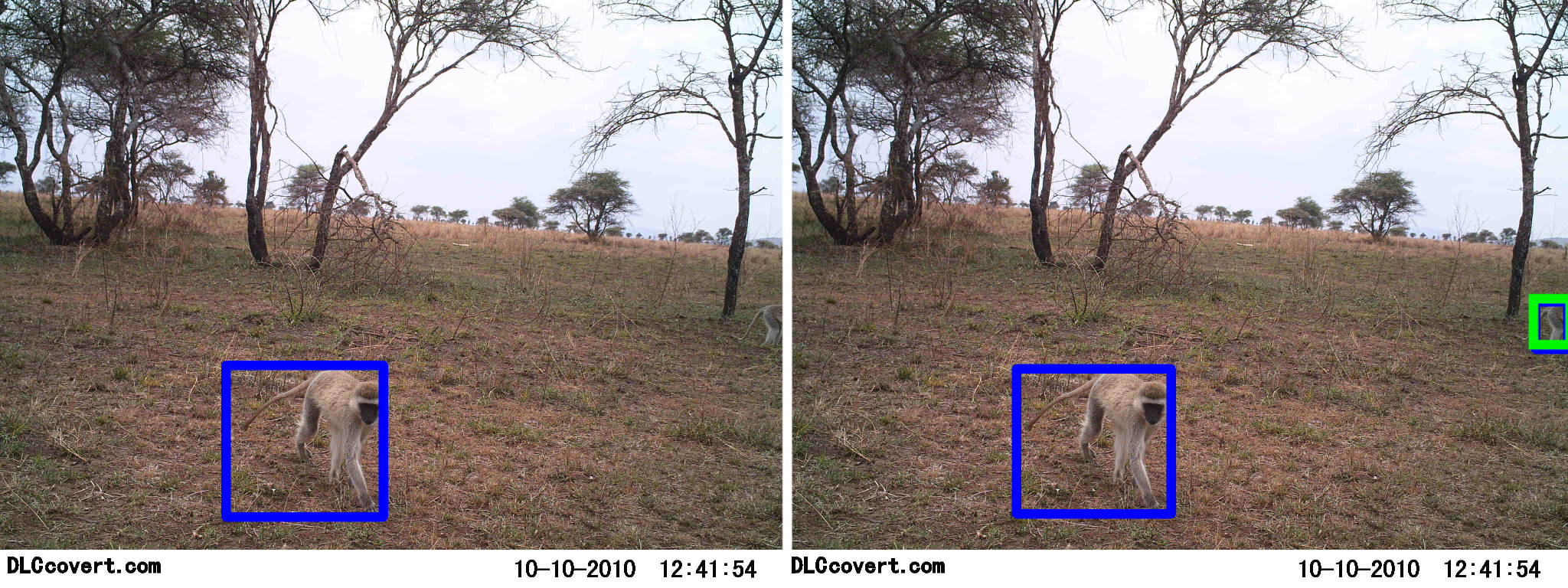}}
\vspace{-10pt} \\

\subfloat[\label{fig:citycam_bad_condition}][Objects poorly lit.]{\includegraphics[width=8cm,height=2.65cm]{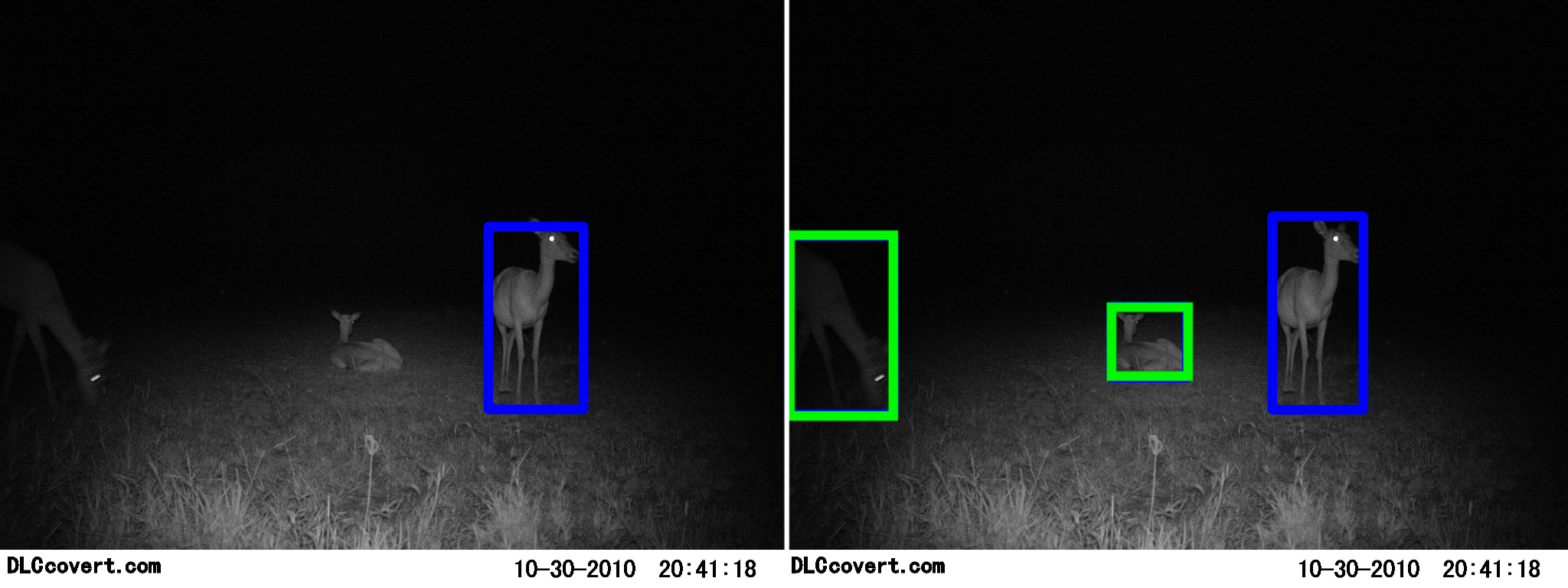}}
\vspace{-10pt} \\

\subfloat[\label{fig:salient_object}][Background distractor.]{\includegraphics[width=8cm,height=2.65cm]{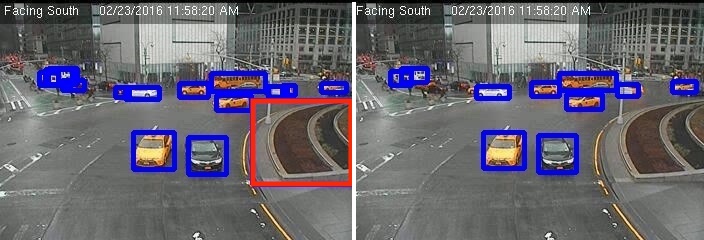}}
\vspace{-2pt} \\

\Caption{\textbf{Static Monitoring Camera Challenges.} Images taken without a human photographer have no quality guarantees; we highlight challenges which cause mistakes in single-frame systems (left) and are fixed by our model (right). False single-frame detections are in \textbf{\textcolor{red}{red}}, detections missed by the single-frame model and corrected by our method are in \textbf{\textcolor{green}{green}}, and detections that are correct in both models are in \textbf{\textcolor{blue}{blue}}. Note that in camera traps, the intra-image context is very powerful due to the group behavior of animal species.}
\vspace{-5pt}
\label{fig:data_challenge}
\end{figure}

These cases are often difficult even for humans. On the other hand, there are aspects of the passive monitoring problem domain that can give us hope --- for example,
subjects often exhibit similar behavior over multiple days, and background objects remain static, suggesting that it would be beneficial to provide temporal context in the  form  of additional frames from  the  same  camera. Indeed we would expect humans viewing passive monitoring footage to often rewind to get better views of a difficult-to-see object. 

These observations forms the intuitive basis for our model that can learn how to find and use other potentially easier examples from the same camera to help improve detection performance (see Figure \ref{fig:data_challenge}). Further, like most real-world data \cite{van2018inaturalist}, both  traffic camera and camera trap data have long-tailed class distributions. By providing context for rare classes from other examples, we improve performance  in the long tail as well as on common classes. 

More specifically, we propose a detection architecture, \emph{Context R-CNN}, that learns to differentiably index into a long-term memory bank while performing detection within a static camera. This architecture
is flexible and is applicable even in the aforementioned low, variable framerate scenarios. 
At a high level, our approach can be framed as a non-parametric estimation method (like nearest neighbors) sitting on top of a high-powered parametric function (Faster R-CNN). When train and test locations are quite different, one might not expect a parametric method to generalize well \cite{beery2018recognition}, whereas Context R-CNN is able to leverage an unlabeled `neighborhood' of test examples for improved generalization. 

\vspace{-2mm}
\paragraph{We focus on two static-camera domains:}\vspace{-1mm}
\begin{itemize}\denselist
\item \textbf{Camera traps} are remote static monitoring cameras used by biologists to study animal species occurrence, populations, and behavior. Monitoring biodiversity quantitatively can help us understand the connections between species decline and pollution, exploitation, urbanization, global warming, and  policy. 
\item \textbf{Traffic cameras} are static monitoring cameras used to monitor roadways and intersections in order to analyze traffic patterns and ensure city safety. 
\end{itemize}\vspace{-1mm}
In both domains, the contextual signal within a single camera location is strong, and we allow the network to determine which previous images were relevant to the current frame, regardless of their distance in the temporal sequence.  This is important within a static camera, as objects exhibit periodic, habitual behavior that causes them to appear days or even weeks apart. For example, an animal might follow the same trail to and from a watering hole in the morning and evening every night, or a bus following its route will return periodically throughout the day. 

\vspace{-2mm}
\paragraph{To summarize our main contributions:}\vspace{-1mm}
\begin{itemize}\denselist
\item We propose \emph{Context R-CNN}, which leverages temporal context for improving object detection regardless of frame rate or sampling irregularity. It can be thought of as a way to improve generalization to novel cameras by incorporating unlabeled images.
\item We demonstrate major improvements over strong single-frame baselines; on a commonly-used camera trap dataset we improve mAP at 0.5 IoU by 17.9\%.% on two camera trap datasets as well as a traffic camera dataset.  
\item We show that Context R-CNN is able to leverage up to a month of temporal context which is significantly more than prior approaches.
\end{itemize}
\vspace{-3mm}

\Section{Related Work}
\noindent
\textbf{Single frame object detection.}
Driven by popular benchmarks such as COCO~\cite{lin2014microsoft} and Open Images \cite{kuznetsova2018open}, there have been a number of advances in single frame object detection in recent years.  These detection architectures include anchor-based models, both single stage (e.g., SSD \cite{liu2016ssd}, RetinaNet \cite{lin2017focal}, Yolo \cite{redmon2016you,redmon2017yolo9000}) and two-stage (e.g., Fast/Faster R-CNN~\cite{girshick2015fast,ren2015faster,he2016deep}, R-FCN~\cite{dai2016r}), as well as more recent anchor-free models (e.g., CornerNet~\cite{law2018cornernet}, CenterNet~\cite{zhou2019objects}, FCOS~\cite{tian2019fcos}). Object detection methods have shown great improvements on COCO- or Imagenet-style images, but these gains do not always generalize to challenging real-world data (See Figure \ref{fig:data_challenge}).

\noindent
\textbf{Video object detection.}
Single frame architectures then form the basis for video detection and spatio-temporal action localization architectures, which build upon single frame models by incorporating contextual cues from other frames in order to deal with more specific challenges that arise in video data including motion blur, occlusion, and rare poses. Leading methods have used pixel level flow (or flow-like concepts) to aggregate features~\cite{zhu2017deep,zhu2017flow,zhu2018towards,bertasius2018object} or used correlation~\cite{feichtenhofer2017detect} to densely relate features at the current timestep to an adjacent timestep.  Other papers have explored the use of 3d convolutions (e.g., I3D, S3D)~\cite{carreira2017quo,luo2018fast,xie2018rethinking} or recurrent networks~\cite{liu2018mobile,kang2017object} to extract better temporal features.  Finally, many works apply video specific postprocessing to “smooth” predictions along time, including tubelet smoothing~\cite{gkioxari2015finding} or SeqNMS~\cite{han2016seq}.

\noindent
\textbf{Object-level attention-based temporal aggregation methods.}
The majority of the above video detection approaches are not well suited to our target setting of sparse, irregular frame rates. For example, flow based methods, 3d convolutions and LSTMs typically  assume a dense, regular temporal sampling.  And while models like LSTMs can theoretically depend on all past frames in a video, their effective temporal receptive field is typically much smaller.  To address this limitation of recurrent  networks, the NLP community has introduced attention-based architectures as a way to take advantage of long range dependencies in sentences~\cite{bahdanau2014neural,vaswani2017attention,devlin2018bert}.  The vision community has followed suit with attention-based architectures~\cite{sun2019videobert,lu2019vilbert,su2019vl} that leverage longer term temporal context. 

Along the same lines and most relevant to our work, there are a few recent works ~\cite{wu2019long,shvets2019leveraging,wu2019sequence,deng2019object} that rely on non-local attention mechanisms in order to aggregate information at the object level across time. For example, Wu et al~\cite{wu2019long} applied non-local attention~\cite{wang2018non} to person detections to accumulate context from pre-computed feature banks (with frozen pre-trained feature extractors).  These feature banks extend the time horizon of their network up to 60s in each direction, achieving strong results on spatio-temporal action localization. 
We similarly use a frozen feature extractor that allows us to create extremely long term memory banks which leverage the spatial consistency of static cameras and habitual behavior of the subjects across long time horizons (up to a month). However Wu et al use a 3d convnet (I3D) for short term features which is not well-suited to our setting due to low, irregular frame rate.  Instead we use a single frame model for the current frame which is more similar to \cite{shvets2019leveraging,wu2019sequence,deng2019object} who proposed variations of this idea for video object detection achieving strong results on the Imagenet Vid dataset.  In contrast to these three papers, we augment our model with an additional dedicated short term attention mechanism which we show to be effective in experiments. Uniquely, our approach also allows negative examples into memory which allows the model to learn to ignore salient false positives in empty frames due to their immobility; we find that our network is able to learn background classes (e.g., rocks, bushes) without supervision.

More generally, our paper adds to the growing evidence that this attention-based approach of temporally aggregating information at the object level is highly effective for incorporating more context in video understanding.  We argue in fact that it is especially useful in our setting of sparse irregular frame samples from static cameras.  Whereas a number of competing baselines like 3d convolutions and flow based techniques perform nearly as well as these attention-based models on Imagenet Vid, the same baselines are not well-suited to our setting. Thus, we see a larger performance boost from prior, non-attention-based methods to our attention-based approach.

\noindent
\textbf{Camera traps and other visual monitoring systems.}
Image classification and object detection have been increasingly explored as a tool for reducing the arduous task of classifying and counting animal species in camera trap data \cite{beery2018recognition, norouzzadeh2018automatically, yousif2017fast, miguel2016finding, villa2017towards, zhang2016animal, yu2013automated, schneider2018deep, beery2019synthetic, beery2019efficient}. Detection has been shown to greatly improve the generalization of these models to new camera locations \cite{beery2018recognition}. It has also been  shown in \cite{beery2018recognition, norouzzadeh2018automatically, yousif2017fast} that temporal information is useful. 
However, previous methods cannot report per-image species identifications (instead identifying a class at the burst level), cannot handle image bursts containing multiple species, and cannot provide per-image localizations and thus species counts, all of which are important to biologists. 

In addition, traffic cameras, security cameras, and weather cameras on mountain passes are all frequently stationary and used to monitor places over long time scales. For traffic cameras, prior work focuses on crowd counting (e.g., counting the number of vehicles or humans in each image) \cite{zhang2017understanding,zhao2018adversarial,arteta2016counting,chan2008privacy,shah2018cadp}. Some recent works have investigated using temporal information in traffic camera datasets \cite{zhang2017fcn,xiong2017spatiotemporal}, but these methods only focus on short term time horizons, and do not take advantage of long term context.

%\vspace{-3pt}
\Section{Method}
\begin{figure*}[ht!]

\subfloat[\label{fig:architecture}][High-level Context R-CNN architecture.]{\includegraphics[width=12.5cm]{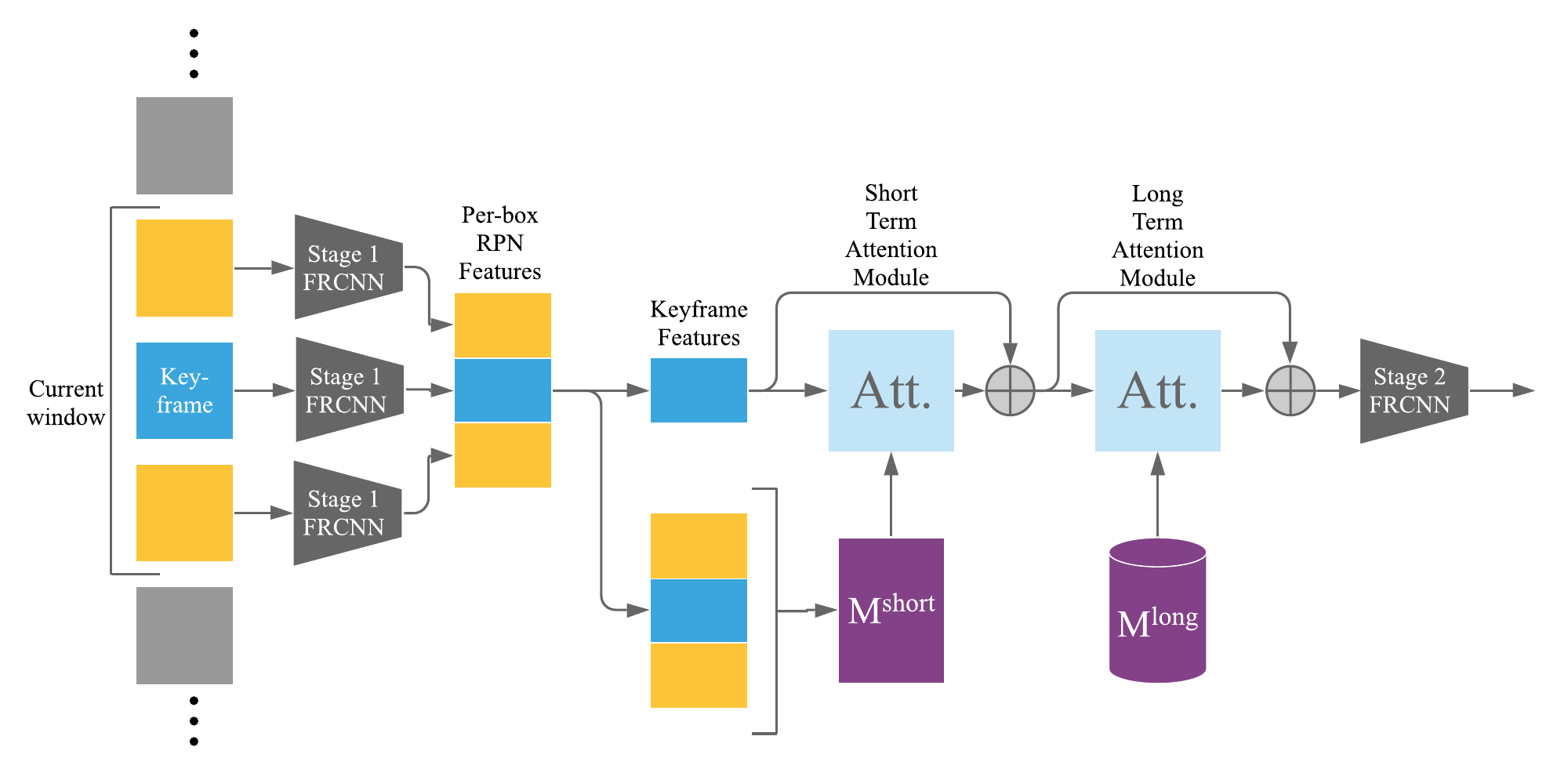}}\hfill
\subfloat[\label{fig:attnblock}][Single attention block.]{\includegraphics[width=4.5cm]{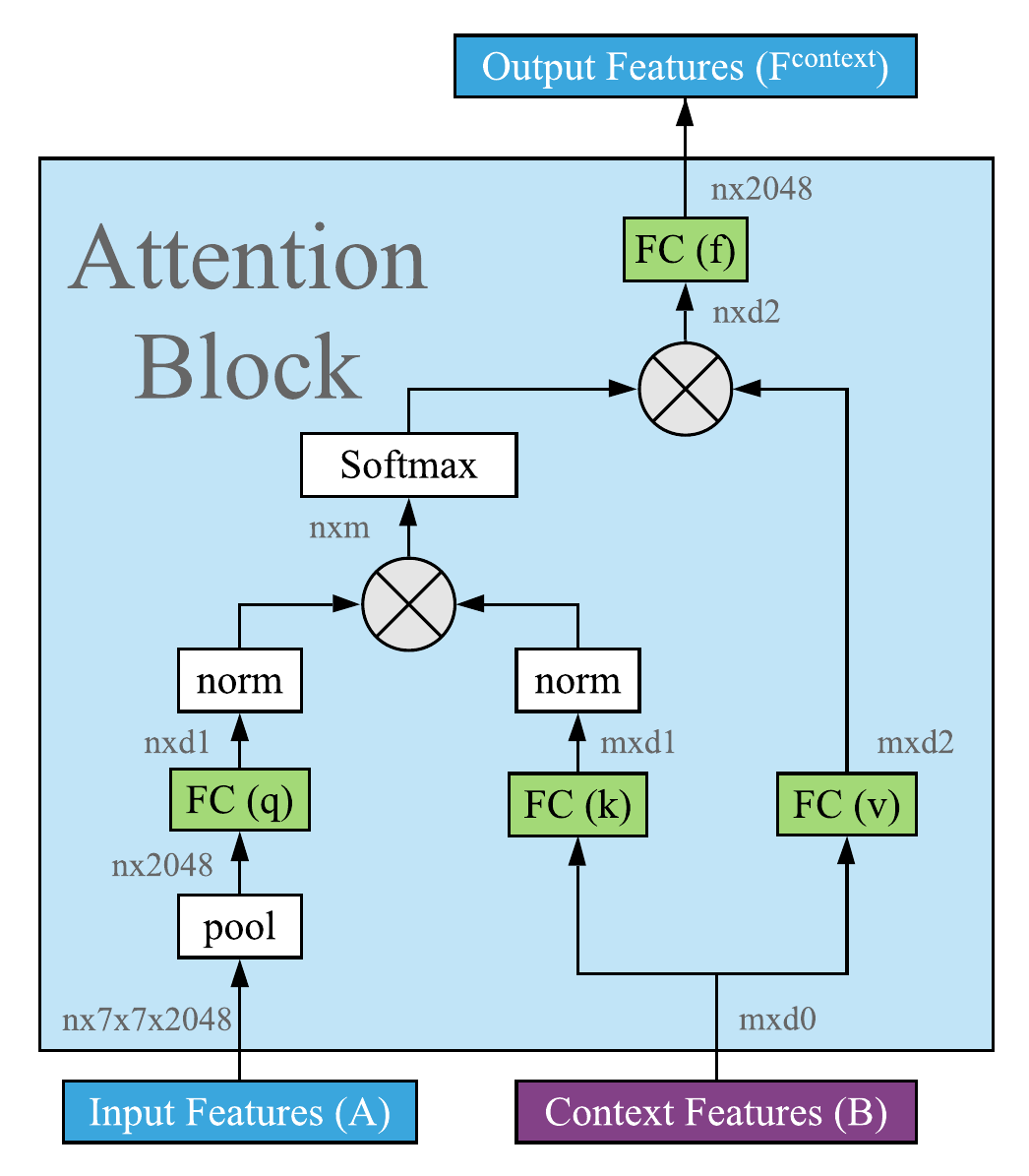}}
\Caption{\textbf{Context R-CNN Architecture. (a)} The high-level architecture of the model, with short term and long term attention used sequentially. Short term and long term attention are modular, and the system can operate with either or both. \textbf{(b)} We see the details of our implementation of an attention block, where  $n$ is the number of boxes proposed by the RPN for the keyframe, and $m$ is the number of comparison features. For short term attention, $m$ is the total number of proposed boxes across all frames in the window, shown in (a) as $M^{short}$. For long term attention, $m$ is the number of features in the long term memory bank $M^{long}$ associated with the current clip. See Section \ref{sec:mem_bank} for details on how this memory bank is constructed.}
% \vspace{-3pt}
\label{fig:all_arch}
\end{figure*}   

Our proposed approach, Context R-CNN, builds a ``memory bank'' based on contextual frames and modifies a detection model to make predictions conditioned on this memory bank. In this section we discuss (1) the rationale behind our choice of detection architecture, (2) how to represent contextual frames, and (3) how to incorporate these contextual frame features into the model to improve current frame predictions.

Due to our sparse, irregular input frame rates, typical temporal architectures such as 3d convnets and recurrent neural networks are not well-suited, due to a lack of inter-frame temporal consistency (there are significant changes between frames). Instead, we build Context R-CNN on top of single frame detection models. Additionally, building on our intuitions that moving objects exhibit periodic behavior and tend to appear in similar locations, we hope to inform our predictions by conditioning on instance level features from contextual frames.  Because of this last requirement, we choose the Faster R-CNN architecture \cite{ren2015faster} as our base detection model as this model remains a highly competitive meta-architecture and provides clear choices for how to extract instance level features. Our method is easily applicable to any two stage detection framework.  %We note that while Faster R-CNN often is viewed as being slower than its single shot competitors (e.g., RetinaNet~\cite{lin2017focal}), our effectively lower frame rates moot the need for real time inference.  Moreover, passive monitoring cameras such as camera traps are traditionally not cloud connected and only analyzed periodically in batch.

As a brief review, Faster R-CNN proceeds in two stages. An image is first passed through a first-stage region proposal network (RPN) which, after running non-max suppression, returns a collection of class agnostic bounding box proposals. These box proposals are then passed into the second stage, which extracts instance-level features via the ROIAlign  operation~\cite{he2017mask,huang2017speed} which then undergo classification and box refinement.

In Context R-CNN, the first-stage box proposals are instead routed through two attention-based modules that (differentiably) index into memory banks, allowing the model to incorporate features from contextual frames (seen by the same camera) in order to provide local and global temporal context. These attention-based modules return a contextually-informed feature vector which is then passed through the second stage of Faster R-CNN in the ordinary way.  In the following section (\ref{sec:mem_bank}), we discuss  how to represent features from context frames using a memory bank and detail our design of the attention modules.  See Figure~\ref{fig:all_arch} for a diagram of our pipeline.

\begin{figure*}[h]
\vspace{-10pt}
\centering
\includegraphics[width=16cm]{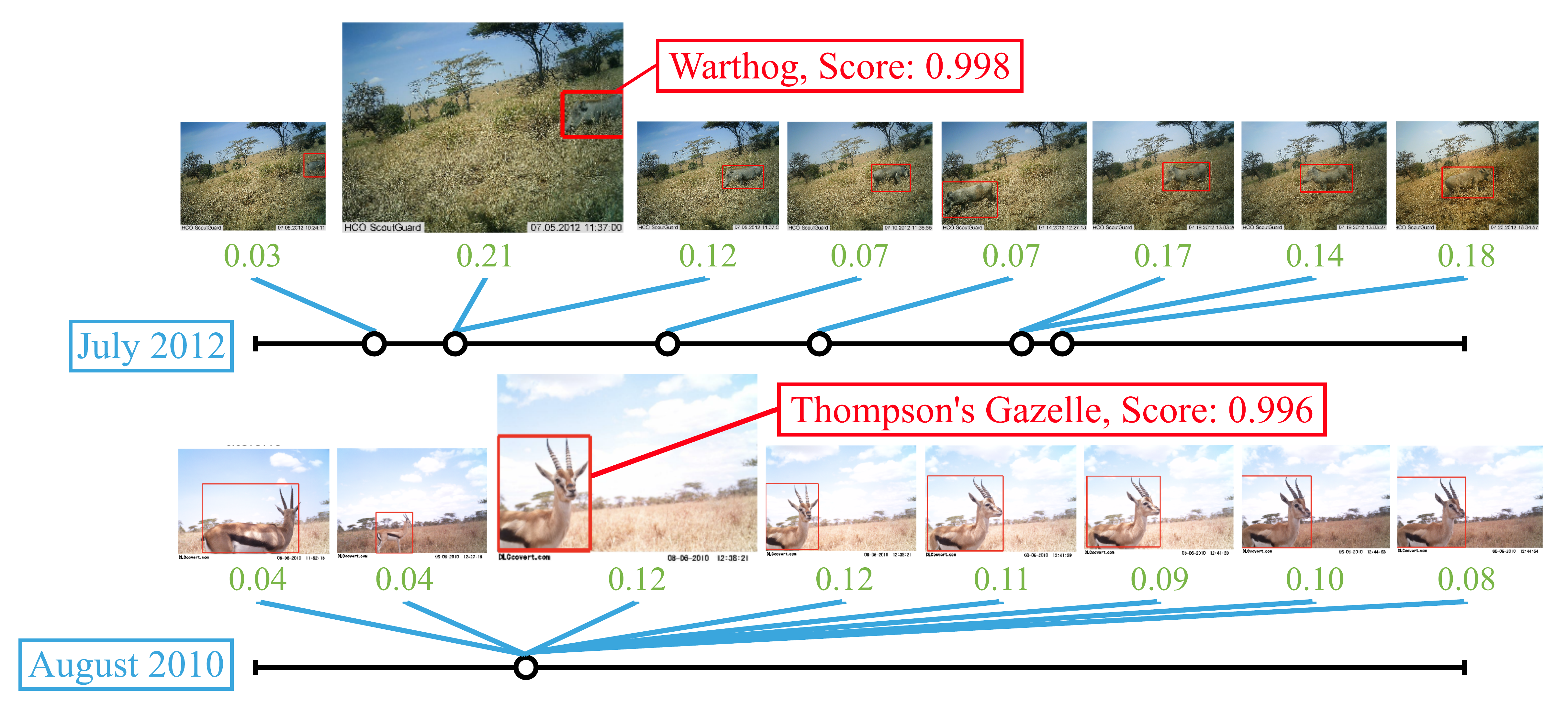}
\Caption{\textbf{Visualizing attention.} In each example, the keyframe is shown at a larger scale, with Context R-CNN's detection, class, and score shown in red. We consider a time horizon of one month, and show the images and boxes with highest attention weights (shown in green). The model pays attention to objects of the same class, and the distribution of attention across time can be seen in the timelines below each example. A warthogs' habitual use of a trail causes useful context to be spread out across the month, whereas a stationary gazelle results in the most useful context to be from the same day. The long term attention module is adaptive, choosing to aggregate information from whichever frames in the time horizon are most useful.}
\label{fig:attention_viz}
% \vspace{-3pt}
\end{figure*}

\Subsection{Building a memory bank from context features}\label{sec:mem_bank}

\noindent
\textbf{Long Term Memory Bank ($M^{long}$)}.
Given a keyframe $i_t$, for which we want to detect objects, we iterate over all frames from the same camera within a pre-defined time horizon $i_{t-k}:i_{t+k}$, running a frozen, pre-trained detector on each frame. We build our long term memory bank ($M^{long}$) from feature vectors corresponding to resulting detections.
Given the limitations of hardware memory, deciding what to store in a memory bank is a critical design choice. We use three strategies to ensure that our memory bank can feasibly be stored.  
\vspace{-5pt}
\begin{itemize}\denselist
    \item We take instance level feature tensors after cropping proposals from the RPN and save only a spatially pooled representation of each such tensor concatenated with a spatiotemporal encoding of the datetime and box position (yielding per-box embedding vectors).
    \item We curate by limiting the number of proposals for which we store features --- we consider multiple strategies for deciding which and how many features to save to our memory banks, see Section \ref{memory_rep} for more details.
    \item We rely on a 
pre-trained single frame Faster R-CNN with Resnet-$101$ backbone as a frozen feature extractor (which therefore need not be considered during backpropagation).  In experiments we consider an extractor pretrained on COCO alone, or fine-tuned on the training set for each dataset. We find that COCO features can be used effectively but that best performance comes from a fine-tuned extractor (see Table \ref{tab:results}(c)).
\end{itemize}
\vspace{-5pt}
Together with our sparse frame rates, by using these strategies we are able to construct memory banks holding up to $8500$ contextual
features --- in our datasets, this is sufficient to represent a month's worth of context from a camera.

% a month of context in memory. This corresponds to anywhere from $1$ to $8500$ contextual features, depending on the number of images taken during the month by the camera (with an average of $74$ contextual frames per keyframe).

\noindent
\textbf{Short Term Memory ($M^{short}$).}
In our experiments we show that it is helpful to include a separate 
mechanism for incorporating short term context features from nearby frames, using the same, trained first-stage feature extractor as for the keyframe. This is different from our long term memory from above which we build over longer time horizons with a frozen feature extractor. In contrast to  long term memory, we do not curate the short term features: for small window sizes it is feasible to hold features for all box proposals in memory.
We take the stacked tensor of cropped instance-level features across all frames within a small window around the current frame (typically $\leq 5$ frames) and globally pool across the spatial dimensions (width and height). 
This results in a matrix of shape $(\mbox{\# proposals per frame} *  \mbox{\# frames}) \times \mbox{(feature depth)}$
containing a single embedding vector per box proposal (which we call our 
\emph{Short Term Memory}, $M^{short}$), that is then passed into the
short term attention block.

\Subsection{Attention module architecture}
We define an attention block \cite{vaswani2017attention} which aggregates from context features keyed by input 
features as follows (see Figure \ref{fig:all_arch}):
 Let $A$ be the tensor of input features from the current frame (which in 
 our setting has shape 
 $[n\times 7 \times 7 \times 2048]$, with $n$ the number of proposals emitted by the
 the first-stage of Faster R-CNN). 
 We first spatially pool $A$ across the feature width and height dimensions, yielding $A^{pool}$ with shape $[n\times 2048]$. 
 Let $B$ be the matrix of context features, which has shape $[m\times d_0]$. 
 We set $B=M^{short}$ or $M^{long}$.
 We define $k(\cdot;\theta)$ as the \textit{key} function, $q(\cdot;\theta)$ as the \textit{query} function, $v(\cdot;\theta)$ as the \textit{value} function, and $f(\cdot;\theta)$ as the final projection that returns us to the correct output feature length to add back into the input features. We use a distinct $\theta$ ($\theta^{long}$ or $\theta^{short}$) for long term or short term attention respectively. In our experiments, $k$, $q$, $v$ and $f$ are all fully-connected layers, with output dimension $2048$. We calculate attention weights $w$ using standard dot-product attention: 
 
\vspace{-5pt}
\begin{equation}
w = \mbox{Softmax}\left((k(A^{pool};\theta) \cdot q(B; \theta)) \,/\,  (T\sqrt{d})\right),\vspace{-1mm}
\label{eqn:attention}
\end{equation}

\noindent
where $T>0$ is the softmax temperature,
$w$ the attention weights  
with shape $[n \times m]$, and $d$ the feature depth ($2048$). 

We next construct a context feature $F^{context}$ for each box by taking a projected, weighted sum of context features:

\vspace{-3pt}
\begin{equation}
F^{context} = f(w \cdot v(B;\theta);\theta),
\label{eqn:aggregation}
\end{equation}

\noindent
where $F^{context}$ has shape $[n \times 2048]$ in our setting.
Finally, we add $F^{context}$ as a per-feature-channel bias back into our original input features $A$.

\Section{Data}

Our model is built for variable, low-frame-rate real-world 
systems of static cameras, and we test our methods on two such domains: camera traps and traffic cameras. 
Because the cameras are static, we split each dataset into separate camera locations for train and test, to ensure our model does not overfit to the validation set \cite{beery2018recognition}. 

\noindent
\textbf{Camera Traps.}
Camera traps are usually programmed to capture an image burst of $1-10$ frames (taken at 1 fps) after each motion trigger, which results in data with variable, low frame rate.  In this paper, we test our systems on the Snapshot Serengeti (SS) \cite{swanson2015snapshot} and Caltech Camera Traps (CCT) \cite{beery2018recognition} datasets, each of which have human-labeled ground truth bounding boxes for a subset of the data. We increase the number of bounding box labeled images for training by pairing class-agnostic detected boxes from the Microsoft AI for Earth MegaDetector \cite{beery2019efficient} with image-level species labels on our training locations. SS has $10$ publicly available seasons of data. We use seasons $1-6$,  containing $225$ cameras, $3.2$M images, and $48$ classes. CCT contains 140 cameras, 243K images, and 18 classes. Both datasets have large numbers of false motion triggers, $75$\% for SS and $50$\% for CCT; thus many images contain no animals.  We split the data using the location splits proposed in \cite{LILA}, and evaluate on the images with human-labeled bounding boxes from the validation locations for each dataset ($64$K images across $45$ locations for SS and $62$K images across $40$ locations for CCT). 
\\
\noindent
\textbf{Traffic Cameras.}
The CityCam dataset \cite{zhang2017understanding} contains $10$ types of vehicle classes, around $60$K frames and $900$K annotated objects. It covers $17$ cameras monitoring downtown intersections and parkways in a high-traffic city, and ``clips'' of data are sampled multiple times per day, across months and years. The data is diverse, covering day and nighttime, rain and snow, and high and low traffic density. We use $13$ camera locations for training and $4$ cameras for testing, with both parkway and downtown locations in both sets.

\begin{table}[t]
\footnotesize
    \subfloat[\label{tab:results_datasets}][Results across datasets]{
        \centering
        \setlength\tabcolsep{6pt}
        \begin{tabular}{r|cc|cc|cc|}
        & \multicolumn{2}{|c|}{SS} & 
        \multicolumn{2}{|c|}{CCT} &
        \multicolumn{2}{|c|}{CC} \\
         Model &mAP&AR&mAP&AR&mAP&AR \\
        \hline
         Single Frame & 37.9 & 46.5 & 56.8 & 53.8 & 38.1 & 28.2\\
         \textbf{Context R-CNN} & \textbf{55.9} & \textbf{58.3} & \textbf{76.3} & \textbf{62.3} & \textbf{42.6} & \textbf{30.2} \\
    \end{tabular}} \\
    \subfloat[\label{tab:time}][Time horizon]{
    \setlength\tabcolsep{4pt}
    \begin{tabular}{r|cc|}
        SS & mAP & AR \\
        \hline
        One minute & 50.3 & 51.4\\
        One hour & 52.1 & 52.5\\
        One day & 52.5 & 52.9\\
        One week & 54.1 & 53.2\\
        \textbf{One month} & \textbf{55.6} & \textbf{57.5}\\
    \end{tabular}
    }\hfill
        \subfloat[\label{tab:sampling}][Selecting memory]{
    \setlength\tabcolsep{4pt} 
    \begin{tabular}{r|cc|}
        SS & mAP & AR  \\
        \hline
        \textbf{One box per frame} & \textbf{55.6} & \textbf{57.5} \\
        COCO features & 50.3 & 55.8\\
        Only positive boxes & 53.9 & 56.2\\
        Subsample half & 52.5 & 56.1\\
        Subsample quarter & 50.8 & 55.0\\
    \end{tabular}
    }\\
    \subfloat[\label{tab:SS_baselines}][Comparison across models]{
        \centering
        \setlength\tabcolsep{4pt} 
        \begin{tabular}{r|cc|}
         SS &mAP&AR \\
        \hline
         Single Frame & 37.9 & 46.5 \\
         \hline
         Maj. Vote & 37.8 & 46.4\\
         ST Spatial & 39.6 & 36.0 \\
         %LSTD &  &  \\
         S3D & 44.7 & 46.0 \\
         \hline
         SF Attn & 44.9 & 50.2 \\
         ST Attn & 46.4 & 55.3 \\
         LT Attn & 55.6 & 57.5 \\
         \textbf{ST+LT Attn} & \textbf{55.9} & \textbf{58.3} \\
    \end{tabular}
    }\hfill   
    \subfloat[\label{tab:add_boxes}][Adding boxes to $M^{long}$]{
    \begin{tabular}{r|cc|}
        CC & mAP & AR \\
        \hline
        Single Frame & 38.1 & 28.2 \\
        \hline
        Top 1 Box &  40.5 & 29.3\\
        \textbf{Top 8 Boxes}& \textbf{42.6} & \textbf{30.2}\\
        % \textbf{top 8 + DO} & \textbf{41.5} & 28.6\\
    \end{tabular}
    }
    \Caption{\textbf{Results.} All results are based on Faster R-CNN with a Resnet 101 backbone. We consider the Snapshot Serengeti (SS), Caltech Camera Traps (CCT), and CityCam (CC) datasets. All mAP values employ an IoU threshold of 0.5, and AR is reported for the top prediction (AR@1).}
    \label{tab:results}
\end{table}

\Section{Experiments}
We evaluate all models on held-out camera locations, using established object detection metrics: mean average precision (mAP) at 0.5 IoU and Average Recall (AR). We compare our results to a (comparable) single-frame baseline for all three datasets. We focus the majority of our experiments on a single dataset, Snapshot Serengeti,  investigating the effects of both short term and long term attention, the feature extractor, the long term time horizon,  and the frame-wise sampling strategy for $M^{long}$. We further explore the addition of multiple features per frame in CityCam.

\Subsection{Main Results}
Context R-CNN strongly outperforms the single-frame Faster RCNN with Resnet-101 baseline on both the Snapshot Serengeti (SS) and Caltech Camera Traps (CCT) datasets, and shows promising improvements on CityCam (CC) traffic camera data as well (See Table \ref{tab:results} (a)). For all experiments, unless otherwise noted, we use a fine-tuned dataset specific feature extractor for the memory bank. \textbf{We show an absolute mAP at 0.5 IoU improvement of $\boldsymbol{19.5}$\% on CCT, $\boldsymbol{17.9}$\% on SS, and $\boldsymbol{4.5}$\% on CC.}  Recall improves as well, with AR@$1$ improving $2$\% on CC, $11.8$\% on SS, and $8.5$\% on CCT. 

\begin{figure}
\vspace{-10pt}
\includegraphics[width=8.5cm]{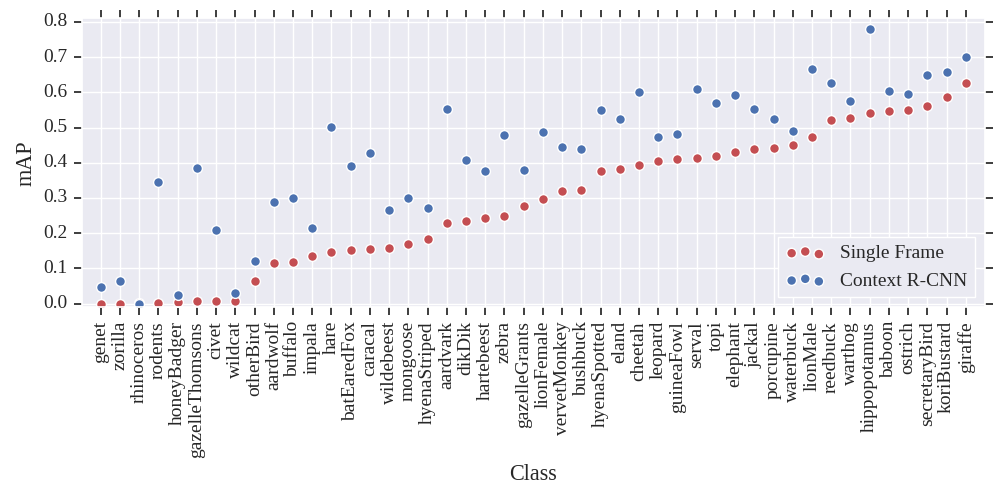}
\vspace{-10pt}
\Caption{\textbf{Performance per class.} Our performance improvement is consistent across classes: we visualize SS per-species mAP from the single-frame model to our best long term and short term memory model.}
\label{fig:performance_per_class}
\vspace{3pt}
\end{figure}

For SS, we also compare against several baselines with access to short term temporal information (Table \ref{tab:results}(d)). All short term experiments use an input window of 3 frames.
\noindent
Our results are as follows:
\begin{itemize}\denselist
   \vspace{-5pt}
    \item We first consider a simple majority vote \textbf{(Maj. Vote)} across the high-confidence single-frame detections within the window, and find that it does not improve over the single-frame baseline.  
    \item We attempt to leverage the static-ness of the camera by taking a temporal-distance-weighted average of the RPN box classifier features from the key frame with the cropped RPN features from the same box locations from the surrounding frames \textbf{(ST Spatial)}, and find it outperforms  the single-frame baseline by $1.9$\% mAP. 
    \item \textbf{S3D} \cite{xie2018rethinking}, a popular video object detection model, outperforms single-frame by $6.8$\% mAP despite being designed for consistently sampled high frame rate video. 
    \item Since animals in camera traps occur in groups, cross-object intra-image context is valuable. An intuitive baseline is to restrict the short term attention context window ($M^{short}$) to the current frame \textbf{(SF Attn)}. This removes temporal context, showing how much improvement we gain from explicitly sharing information across the box proposals in a non-local way. We see that we can gain $7$\% mAP over a vanilla single-frame model by adding this non-local attention module.  
    \item When we increase the short term context window to three frames, keyframe plus two adjacent, \textbf{(ST Attn)} we see an additional improvement of $1.5$\% mAP.  
    \item If we consider \textit{only} long term attention with a time horizon of one month \textbf{(LT Attn)}, we see a $9.2$\% mAP improvement over short term attention.  
    \item By combining both attention modules into a single model \textbf{(ST+LT Attn)}, we see our highest performance at $55.9\%$ mAP, and show in Figure \ref{fig:performance_per_class} that we improve for all classes in the imbalanced dataset.
\end{itemize}

\vspace{-2pt}
\Subsection{Changing the Time Horizon (Table \ref{tab:results}(b))}\label{memory_rep}
We ablate our long term only attention experiments by increasing the time horizon of $M^{long}$, and find that performance increases as the the time horizon increases.
We see a large performance improvement over the single-frame model even when only storing a minute-worth of representations in memory. This is due to the sampling strategy, as highly-relevant bursts of images are
captured for each motion trigger. 
The long term attention block can adaptively determine how to aggregate this information, and there is much useful context across images within a single burst. However, some cameras take only a single image at a trigger; in these cases the long term context becomes even more important. The adaptability of Context R-CNN to be trained on and improve performance across data with not only variable frame rates, but also with different sampling strategies (time lapse, motion trigger, heat trigger, and bursts of $1$-$10$ images per trigger) is a valuable attribute of our system. 

\begin{figure}
\vspace{-10pt}
\subfloat[][Hour]{\includegraphics[width=4cm]{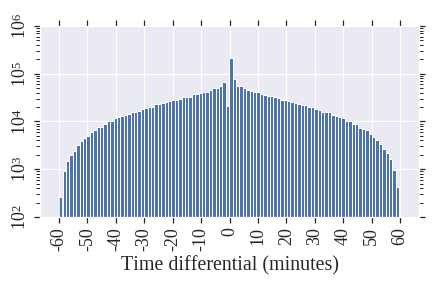}}\hfill
\subfloat[][Day]{\includegraphics[width=4cm]{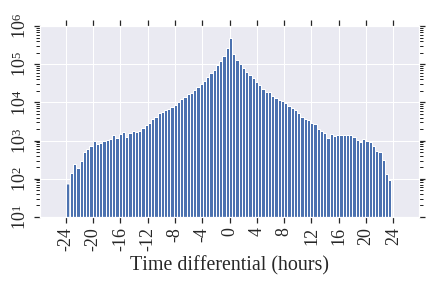}}
\vspace{-12pt}
\\
\subfloat[][Week]{\includegraphics[width=4cm]{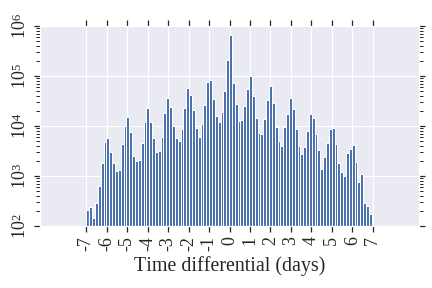}}\hfill
\subfloat[][Month]{\includegraphics[width=4cm]{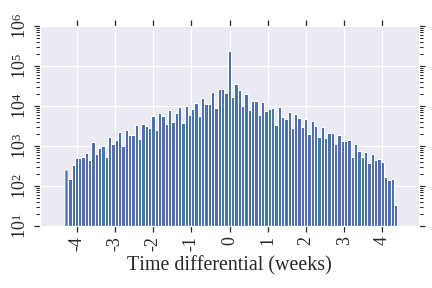}}
\centering
\Caption{\textbf{Attention over time.} 
We threshold attention weights at $0.01$, and plot a histogram of time differentials from the highest-scoring object in the keyframe to the attended frames for varied long term time horizons. Note that the y-axis is in log scale. The central peak of each histogram shows the value of nearby frames, but attention covers the breadth of what is provided: namely, \textbf{if given a month worth of context, Context R-CNN will use it}. Also note a strong day/night periodicity when using a week-long or month-long memory bank.}
\label{fig:attention_hist}
\vspace{3pt}
\end{figure}

In Figure \ref{fig:attention_hist}, we explore the time differential between the top scoring box for each image and the features it most closely attended to, using a threshold of $0.01$ on the attention weight. We can see day/night periodicity in the week- and month-long plots, showing that attention is focused on objects captured at the same time of day. As the time horizon increases, the temporal diversity of the attention module increases and we see that Context R-CNN attends to what is available across the time horizon, with a tendency to focus higher on images nearby in time (see examples in Figure \ref{fig:attention_viz}).
\vspace{-5pt}
\Subsection{Contextual features for constructing $\boldsymbol M^{long}$.}
\noindent
\textbf{Feature extractor (Table \ref{tab:results}(c)).}\label{sec:coco_features}
For Snapshot Serengeti, we consider both a feature extractor trained on COCO, and one trained on COCO and then fine-tuned on the SS training set.  We find that while a month of context from a feature extractor tuned for SS achieves $5.3$\% higher mAP than one trained only on COCO, we are able to outperform the single-frame model by $12.4$\% using memory features that have never before seen a camera trap image. 

\noindent
\textbf{Subsampling memory (Table \ref{tab:results}(c)).}
We further ablate our long term memory by decreasing the stride at which we store representations in the memory bank, while maintaining a time horizon of one month. If we use a stride of $2$, which subsamples the memory bank by half, we see a drop in performance of $3.1$\% mAP at $0.5$. If we increase the stride to $4$, we see an additional $1.7$\% drop. If instead of increasing the stride, we instead subsample by taking only positive examples (using an oracle to determine which images contain animals for the sake of the experiment), we find that performance still drops (explored below). 

\noindent
\textbf{Keeping representations from empty images.}
In our static camera scenario, we choose to add features into our long term memory bank from all frames, both empty and non-empty. The intuition behind this decision is the existence of salient background objects in the static camera frame which do not move over time, and can be repeatedly and erroneously detected by single-frame architectures. We assume that the features from the frozen extractor are visually representative, and thus sufficient for both foreground and background representation. By saving representations of highly-salient background objects, we thus hope to allow the model to learn per-camera salient background classes and positions without supervision, and to suppress these objects in the detection output.  

\begin{figure}
\vspace{-10pt}
\centering
\includegraphics[width=8cm]{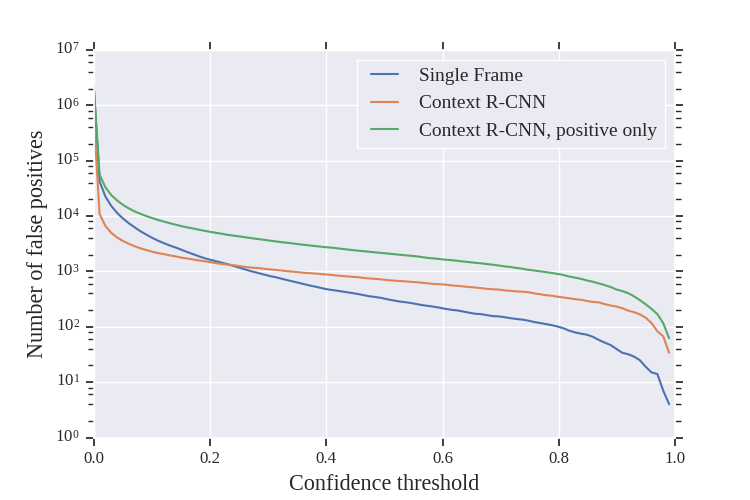}
\Caption{\textbf{False positives on empty images.} When adding features from empty images to the memory bank, we reduce false positives across all confidence thresholds compared to the same model without negative representations. Note that the y-axis is in log scale. The single frame model has fewer high-confidence false positives than either context model, but when given positive and negative context Context R-CNN is able to suppress low-confidence detections. By analyzing Context R-CNN's $100$ most high-confidence detections on images labeled ``empty" we found $97$ images where the annotators missed animals.}
\label{fig:empty_fps}
\vspace{-1pt}
\end{figure}

In Figure \ref{fig:empty_fps}, we see that adding empty representations reduces the number of false positives across all confidence thresholds compared to the same model with only positive representations. We investigated the $100$ highest confidence ``false positives'' from Context R-CNN, and found that in almost all of them ($97/100$), the model had correctly found and classified animals that were missed by  human annotators. The Snapshot Serengeti dataset reports $5$\% noise in their labels \cite{swanson2015snapshot}, and looking at the high-confidence predictions of Context R-CNN on images labeled ``empty'' is intuitively a good way to catch these missing labels. Some of these are truly challenging, where the animal is difficult to spot and the annotator mistake is unfortunate but reasonable. Most are truly just label noise, where the existence of an animal is  obvious, suggesting that our performance improvement estimates are likely conservative.

\noindent
\textbf{Keeping multiple representations per image (Table \ref{tab:results}(e)).}
In Snapshot Serengeti, there are on average $1.6$ objects and $1.01$ classes per image across the non-empty images, and $75\%$ of the images are empty. The majority of the images contain a single object, while a few have large herds of a single species. Given this, choosing only the top-scoring detection to add to memory makes sense, as that object is likely to be representative of the other objects in the image (\textit{e.g.,} keeping only one zebra example from an image with a herd of zebra). In CityCam, however, on average there are $14$ objects and $4$ classes per frame, and only $0.3\%$ of frames are empty. In this scenario, storing additional objects in memory is intuitively useful, to ensure that the memory bank is representative of the camera location. We investigate adding features from the top-scoring $1$ and $8$ detections, and find that selecting $8$ objects per frame yields the best performance (see Table \ref{tab:results}(e)). A logical extension of our approach would be selecting objects to store based not only on confidence, but also diversity.

\noindent
\textbf{Failure modes.} 
One potential failure case of this similarity-based attention approach is the opportunity for hallucination. If one image in a test location contains something that is very strongly misclassified, that one mistake may negatively influence other detections at that camera. For example, when exploring the confident ``false positives'' on the Snapshot Serengeti dataset (which proved to be almost universally true detections that were missed by human annotators) the $3/100$ images where Context R-CNN erroneously detected an animal were all of the same tree, highly confidently predicted to be a giraffe.

% \vspace{-3pt}
\Section{Conclusions and Future Work}
In this work, we contribute a model that leverages per-camera temporal context up to a month, far beyond the time horizon of previous approaches, and show that in the static camera setting, attention-based temporal context is particularly beneficial. Our method, Context R-CNN, is general across static camera domains, improving detection performance over single-frame baselines on both camera trap and traffic camera data. Additionally, Context R-CNN is adaptive and robust to passive-monitoring sampling strategies that provide data streams with low, irregular frame rates.

It is apparent from our results that what and how much information is stored in memory is both important and  domain specific. We plan to explore this in detail in the future, and hope to develop methods for curating diverse memory banks which are optimized for accuracy and size, to reduce the computational and storage overheads at training and inference time while maintaining performance gains.
\vspace{-5pt}

\Section{Acknowlegdements}
We would like to thank Pietro Perona, David Ross, Zhichao Lu, Ting Yu, Tanya Birch and the Wildlife Insights Team, Joe Marino, and Oisin MacAodha for their valuable insight. This work was supported by NSFGRFP Grant No. 1745301, the views are those of the authors and do not necessarily reflect the views of the NSF.

{\small
\bibliographystyle{ieee_fullname}
\bibliography{main}
}

\clearpage
\appendix
\noindent
{\huge Supplementary Material}
\Section{Implementation Details}
We implemented our attention modules within the Tensorflow Object Detection API open-source Faster-RCNN architecture with Resnet 101 backbone \cite{huang2017speed}. Faster-RCNN optimization and model parameters are not changed between the single-frame baseline and our experiments, and we ensure robust single-frame baselines via hyperparameter sweeps. We train on Google TPUs (v3) \cite{kumar2019scale} using MomentumSGD  with  weight  decay  $0.0004$  and  momentum  $0.9$. We construct each batch using $32$ clips, drawing four frames for each clip spaced $1$ frame apart and resizing to $640\times640$. Batches are placed on 8 TPU cores, colocating frames from the same clip. We augment with random flipping, ensuring that the memory banks are flipped to match the current frames to preserve spatial consistency. All our experiments use a softmax temperature of $T=.01$ for the attention mechanism, which we found in early experiments to outperform $.1$ and $1$.

\Section{Dataset Statistics and Per-Class Performance}
Each of the real-world datasets (Snapshot Serengeti, Caltech Camera Traps, and CityCam) has a long-tailed distribution of classes, which can be seen in Figure \ref{fig:ims_per_cat}.  Dealing with imbalanced data is a known challenge across machine learning disciplines \cite{van2018inaturalist, beery2019synthetic}, with rare classes (classes not well-represented during training) frequently proving difficult to recognize.

\begin{figure}[ht!]
  \centering

\subfloat[][Caltech Camera Traps.]{\includegraphics[width=8cm]{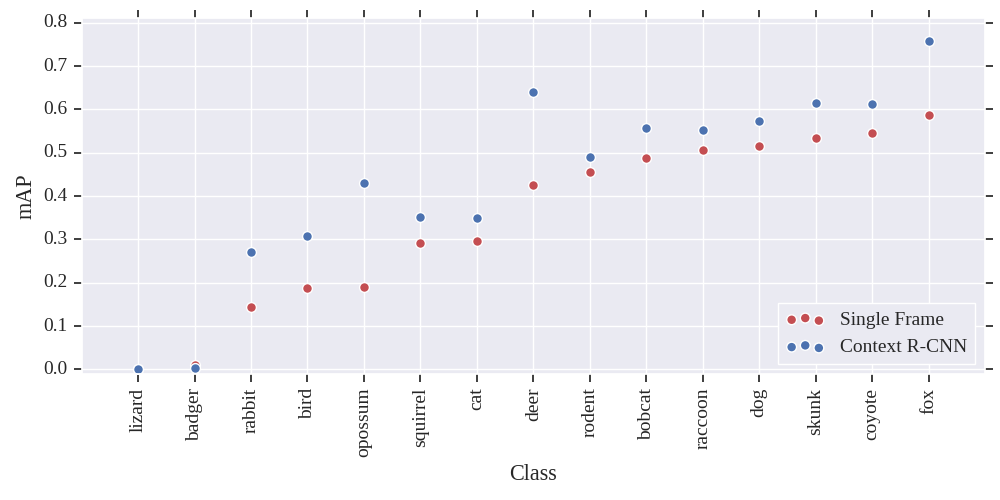}} \\

\subfloat[][CityCam.]{\includegraphics[width=8cm]{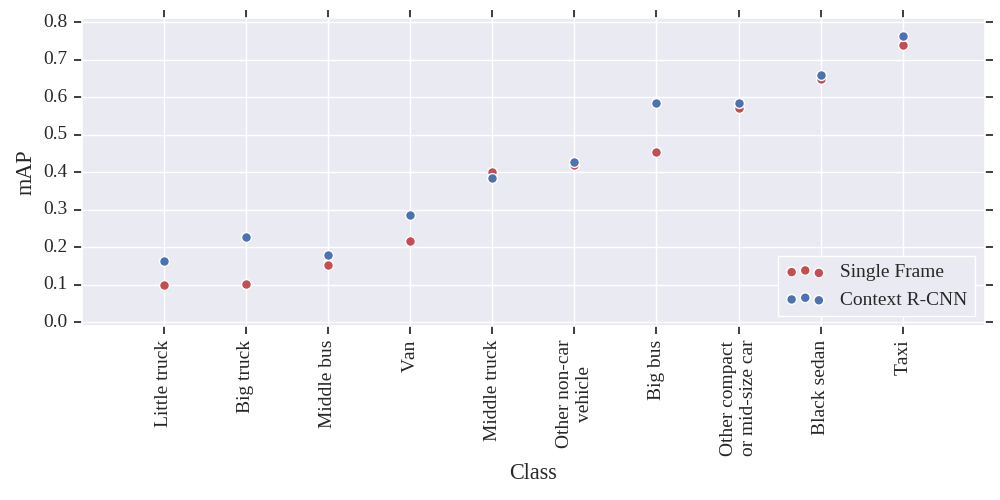}} \\
\Caption{\textbf{Performance per class.} Performance comparison from single-frame to our memory-based model. Note this reports mAP for each class averaged across IoU thresholds, as popularized by the COCO challenge \cite{lin2014microsoft}.}

\label{fig:per_class}
\end{figure}

In Figure 5 in the main text, we demonstrate that the per-class performance universally improves for Snapshot Serengeti (SS).  In Figure \ref{fig:per_class}, we show the per-class performance for Caltech Camera Traps (CCT). and CityCam (CC). Performance on CCT improves for all classes from the single frame model.  We see that for one class in CC, ``Middle Truck", our method performs slightly worse; However, this class is relatively ambiguous, as the concept of ``middle" size is not well-defined.

\begin{figure}
\centering
\subfloat[][Before.]{\includegraphics[width=6cm]{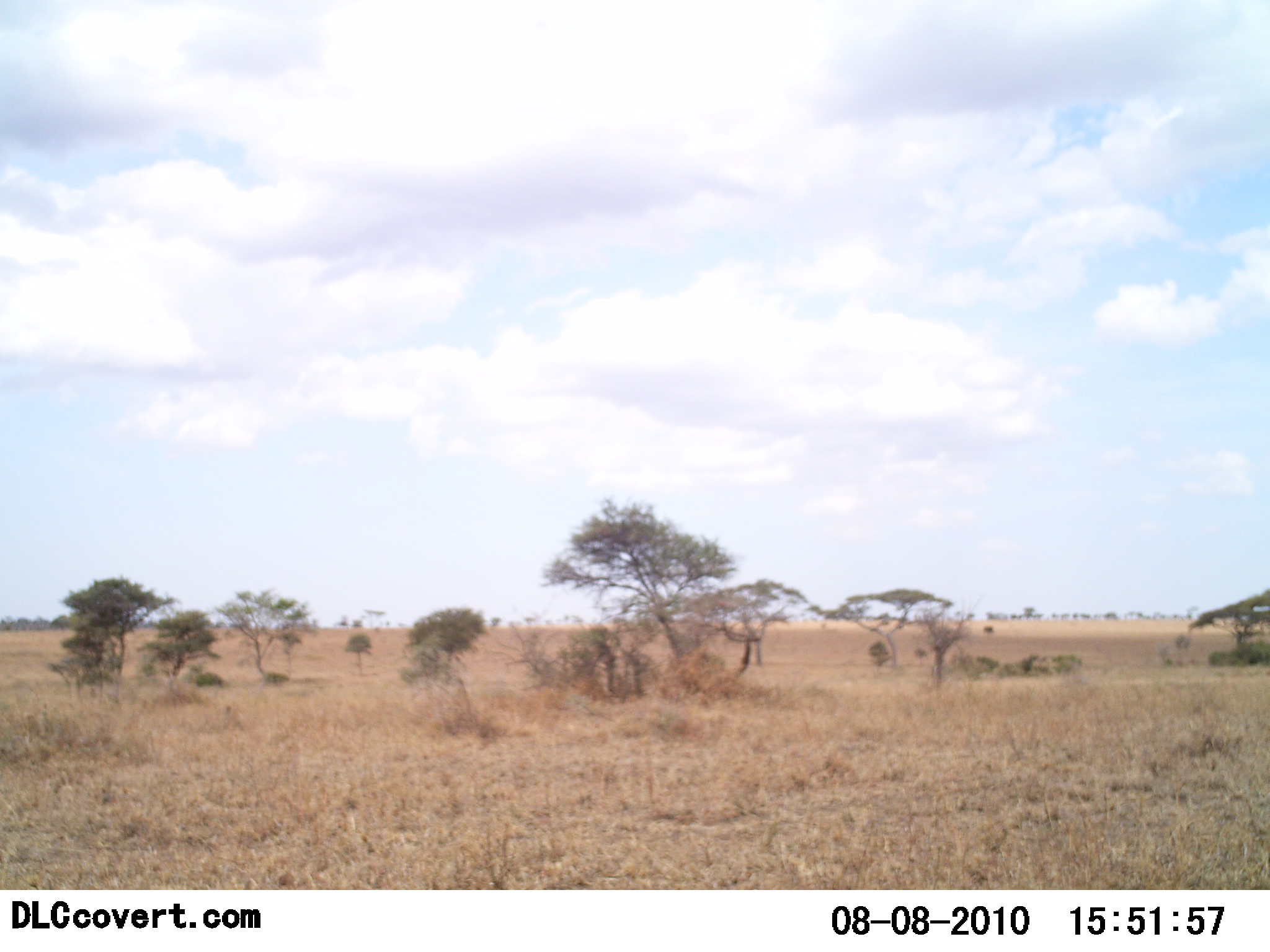}} \hfill

\subfloat[][After.]{\includegraphics[width=6cm]{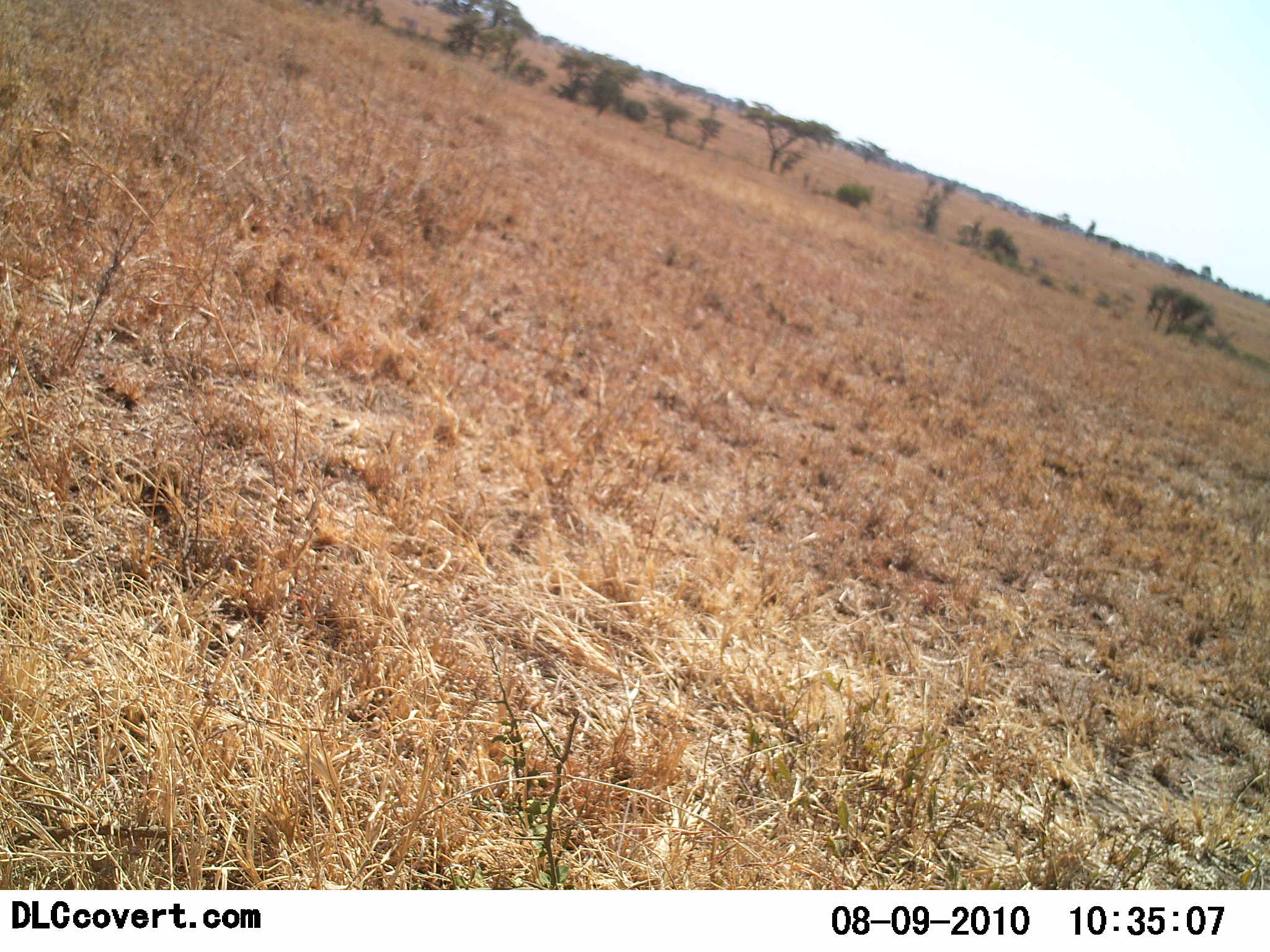}} \\
\Caption{\textbf{Our system is robust to a static camera being accidentally shifted.} Before and after example of a camera that had been bumped by an animal. The images are from the same camera. The first image was taken August 8th 2010, the next August 9th 2010. We find that the system can still utilize contextual information across a camera shift.}

\label{fig:cam_movement}
\end{figure}

\begin{figure}
  \centering

\subfloat[][Snapshot Serengeti.]{
\includegraphics[width=8cm]{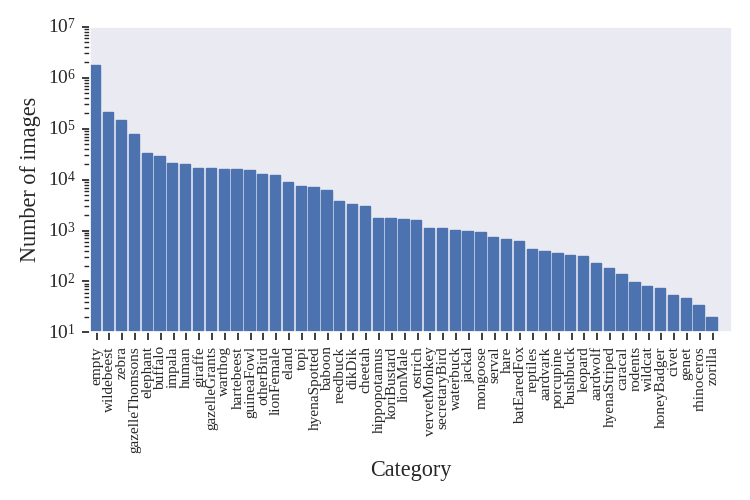}
} \\

\subfloat[][Caltech Camera Traps.]{\includegraphics[width=8cm]{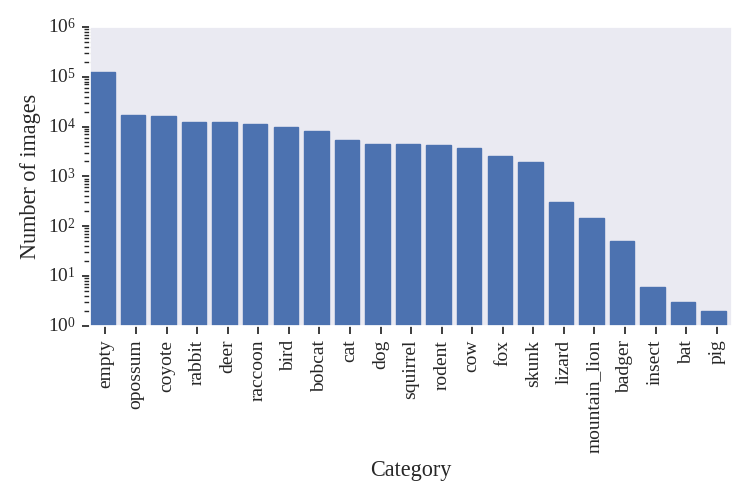}} \\

\subfloat[][CityCam.]{\includegraphics[width=8cm]{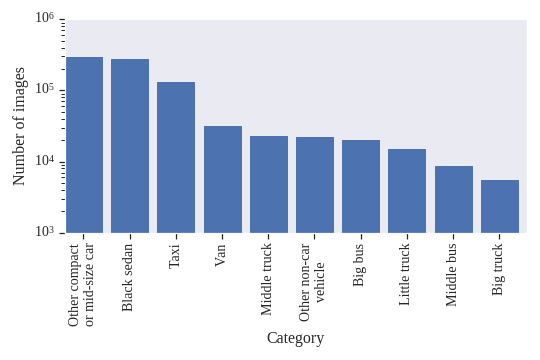}} \\
\Caption{\textbf{Imbalanced class distributions.} Images per category for each of the three datasets. Note the y-axis is in log scale.}

\label{fig:ims_per_cat}
\end{figure}

\Section{Spatiotemptoral Encodings}
We normalize the spatial and temporal information for each object we include in the contextual memory bank. In order to do so, we choose to use a single float between 0 and 1 to represent each of: year, month, day, hour, minute, x center coordinate, y center coordinate, object width, and object height. 

\noindent
We normalize each element as follows:
\begin{itemize}\denselist
    \item \textbf{Year}: We select a reasonable window of possible years covered by our data, 1990-2030. We normalize the year within that window, representing the year in question as $\frac{year-1990}{2030-1990}$.
    \item \textbf{Month}: We normalize the month of the year by 12 months, \textit{i.e.} $\frac{month}{12}$.
    \item \textbf{Day}: We normalize the day of the month by 31 days for simplicity, regardless of how many days there are in the month in question, \textit{i.e.} $\frac{day}{31}$.
    \item \textbf{Hour}: We normalize the hour of the day by 24 hours, \textit{i.e.} $\frac{hour}{24}$.
    \item \textbf{Minute}: We normalize the minute of the hour by 60 minutes, \textit{i.e.} $\frac{minute}{60}$.
    \item \textbf{X Center Coordinate}: We normalize the x coordinate pixel location by the width of the image in pixels, \textit{i.e.} $\frac{x\_center\_location\ (pixels)}{image\_width\ (pixels)}$
    \item \textbf{Y Center Coordinate}: We normalize the y coordinate pixel location by the height of the image in pixels, \textit{i.e.} $\frac{y\_center\_location\ (pixels)}{image\_height\ (pixels)}$
    \item \textbf{Width of Object}: We normalize the object width in pixels by the width of the image in pixels, \textit{i.e.} $\frac{object\_width\ (pixels)}{image\_width\ (pixels)}$
    \item \textbf{Height of Object}: We normalize the object height in pixels by the height of the image in pixels, \textit{i.e.} $\frac{object\_height\ (pixels)}{image\_height\ (pixels)}$
\end{itemize}

\begin{figure*}[h]
\vspace{-10pt}
\centering
\includegraphics[width=16cm]{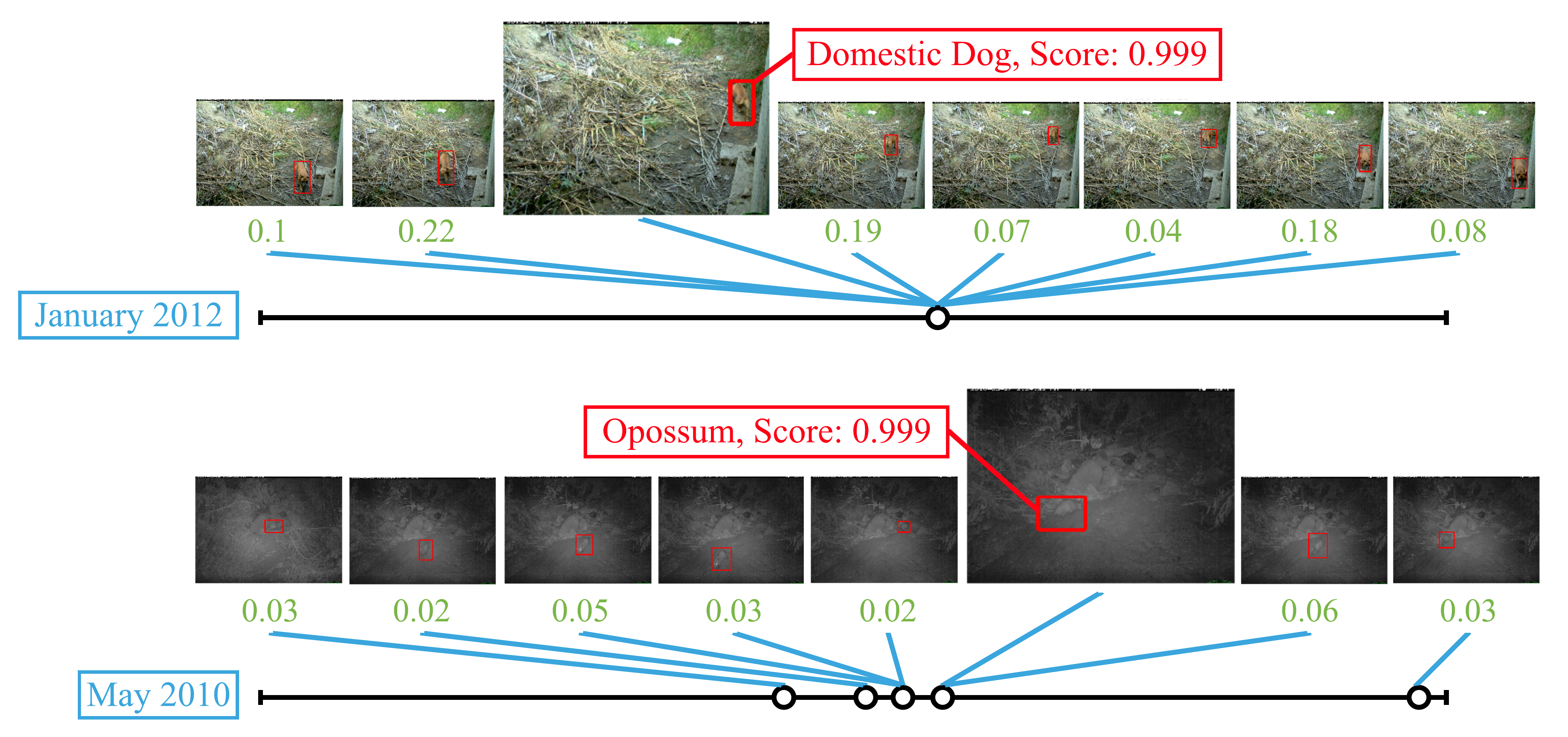}
\Caption{\textbf{Visualizing attention.} In each example, the keyframe is shown at a larger scale, with Context R-CNN's detection, class, and score shown in red. We consider a time horizon of one month, and show the images and boxes with highest attention weights (shown in green). The model pays attention to objects of the same class, and the distribution of attention across time can be seen in the timelines below each example.}
\label{fig:viz}
\vspace{-12pt}
\end{figure*}

\Section{Camera Movement}
Our system has no hard requirements about the camera being static, instead we leverage the fact that it is static implicitly through our memory bank to provide appropriate and relevant context. We find that our system is robust to static cameras that get moved, unlike traditional background modeling approaches.  In Snapshot Serengeti in particular, the animals have a tendency to rub against the camera posts and cause camera shifts over time.  Figure \ref{fig:cam_movement} shows a ``before and after'' 
example of a camera being bumped or moved.

\Section{Attention Visualization}
In Figure 4 in the main text, we visualize attention over time for two examples from Snapshot Serengeti. In Figure \ref{fig:viz} we show examples from Caltech Camera Traps. %the other datasets, CCT and CC. 
Similarly to the visualizations of attention on SS, we see that attention is adaptive to the most relevant information, paying attention across time as needed.  The model consistently learns to attend to objects of the same class. 

\begin{figure*}[h]
\vspace{-10pt}
\centering
\includegraphics[width=16cm]{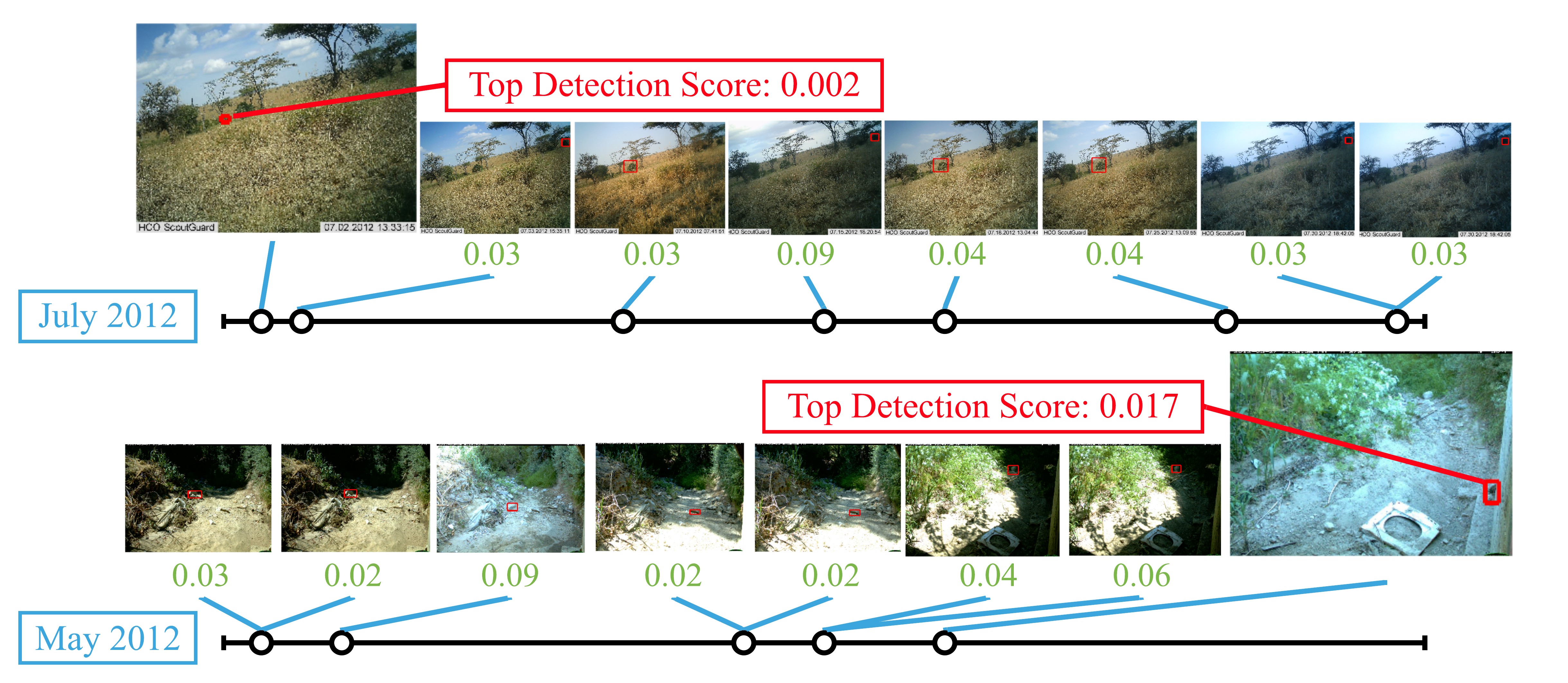}
\Caption{\textbf{Visualizing attention on background classes.} In each example, the keyframe is shown at a larger scale, with Context R-CNN's detection, class, and score shown in red. We consider a time horizon of one month, and show the images and boxes with highest attention weights (shown in green). The first example is from SS, it shows a detected bush (an unlabeled, background class), and shows that Context R-CNN attends to the same bush over time, as well as \textit{ different} bushes in the frame. In the second example, from CCT, we see a similar situation with the background class ``rock."}
\label{fig:viz_empty}
\vspace{-12pt}
\end{figure*}

In Figure \ref{fig:viz_empty}, we visualize how Context R-CNN learns to learn and attend to unlabeled background classes, namely rocks and bushes. Remember that these exact camera locations were never seen during training, so the model has learned to use temporal context to determine when to ignore these salient background classes. It learns to cluster background objects of a certain type, for example bushes, across the frames at a given location. Note that these attended background objects are not always the same instance of the class, which makes sense as background classes may maintain visual similarity within a scene even if they aren't the exact same instance of that type. Species of plants or types of rock are often geographically clustered. 

\end{document}